# Grammar Accuracy Evaluation (GAE): Quantifiable Qualitative Evaluation of Machine Translation Models

**Dojun Park**[1], **Youngjin Jang**[2] and **Harksoo Kim**[2]

[1]Institute of Natural Language Processing, University of Stuttgart

[2]Natural Language Processing Lab, Konkuk University

`dojun.park@ims.uni-stuttart.de, {danyon,nlpdrkim}@konkuk.ac.kr`

**Abstract**

Natural Language Generation (NLG) refers to the operation of expressing the calculation results of a system in human language. Since the quality of generated sentences from an NLG model cannot be fully represented using only quantitative evaluation, they are evaluated using qualitative evaluation by humans in which the meaning or grammar of a sentence is scored according to a subjective criterion. Nevertheless, the existing evaluation methods have a problem as a large score deviation occurs depending on the criteria of evaluators. In this paper, we propose Grammar Accuracy Evaluation (GAE) that can provide the specific evaluating criteria. As a result of analyzing the quality of machine translation by BLEU and GAE, it was confirmed that the BLEU score does not represent the absolute performance of machine translation models and GAE compensates for the shortcomings of BLEU with flexible evaluation of alternative synonyms and changes in sentence structure.

## 1 Introduction

Since language is a product formed over a long period of time based on different geographical and cultural environments, translating meanings between languages into text is a high-level task that requires professional training and high intelligence. With the development of computer processing power, today's machine translation is replacing human labor in various fields of daily life, for instance, SNS such as 'Facebook' and 'Instagram', messenger such as 'Kakao Talk' and 'Line', and 'Google Translate' and 'Microsoft Translate', which translate web pages in conjunction with a web browser.

In machine translation model development, the performance measurement of the model is an important indicator of the progress of the development process. The performance of the machine translation model is measured based on the natural language sentences generated by the model, and the measurement methods are largely divided into qualitative evaluation and quantitative evaluation. Qualitative evaluation is a method of evaluation by a human, and it is considered the most ideal method as a human directly evaluates the translation result generated in human language, but it requires a lot of time and resources and has the disadvantage that the evaluator's personal language bias and subjectivity are reflected. Quantitative evaluation, on the other hand, is a method evaluated by an automated algorithm, and has the advantage of being measured by objective evaluation criteria and being processed quickly by a computer without human effort. However, due to human linguistic characteristics that allow variations of vocabulary and sentence structure, quantitative evaluation that evaluates only based on fixed correct sentences has the limitation of performance measurement, because they cannot distinguish between different forms of sentences that contain the same meaning.

This work proposes Grammar Accuracy Evaluation (GAE), a qualitative evaluation that supplements the limitations of BLEU, which is the most widely used quantitative evaluation for measuring the performance of machine translation models today. First, for the experiment, four machine translation models are trained: Korean-English, English-Korean, German-English, and English-German. Next, the model of the epoch point with the highest BLEU score is selected, and the performance of the model is measured by the GAE to cross-validate the scores between the two evaluation methods. In this way, the significance of the model performance represented by the BLEU score is verified, and the grammatical completeness of each model represented by the GAE score is compared for each category.

## 2 Related Work

### 2.1 BLEU

$$P_n = \frac{\sum_{C \in \{Candidates\}} \sum_{n-gram \in C} Count_{clip}(n-gram)}{\sum_{C' \in \{Candidates\}} \sum_{n-gram' \in C'} Count(n-gram')} \quad (1)$$

$$\begin{cases} BP = 1 & if(c>r) \\ \quad\quad e^{(1-r/c)} & if(c \leq r) \end{cases} \quad (2)$$

$$BLEU = BP \cdot \exp\left(\sum_{n=1}^{N} \frac{1}{N} \log P_n\right) \quad (3)$$

BLEU[1] is one of the quantitative evaluations for measuring the performance of machine translation models and has been the most used evaluation for machine translation models since it appeared in 2002.

The BLEU score is calculated by comparing the output sentence with the reference sentence based on the modified precision (Equation 1). Precision has a problem in that even if the same word is repeated, it is evaluated as a correct translation when it appears more than once in a reference sentence. The modified precision $P_n$ applied to BLEU reflects the adjusted value by limiting the number of occurrences of a word in the translation sentence to the maximum number of occurrences of each word in the reference sentence.

BLEU (Equation 3) is divided into a Brevity Penalty and an N-gram Overlap. The Brevity Penalty (Equation 2) serves to apply a penalty when the length of the translation sentence is too short compared to the reference sentence based on the exponential function, and N-gram overlap is calculated by measuring the degree to which N-grams of the reference sentence (i = 1,...,4) match the N-grams of the translation sentence.

### 2.2 Human Evaluation

[2] measured the performance of machine translation models by applying a qualitative evaluation method that measures adequacy and fluency. Adequacy indicates how well the meaning of the source sentence is expressed in the target sentence of the machine translation model and fluency is evaluated comprehensively by the grammar, spelling, word choice, and style of the translated sentence regardless of the starting sentence. However, it cannot be excluded that the evaluation score measured between 1 and 5 may be relative according to the personal criteria of the evaluator.

[3] used a ranking scale to determine the better result among two or more translation results, given that the quality of the translation is difficult to quantify. In addition, the same score can be given to cases where it is difficult to determine comparative advantage. This method has the advantage of being able to make a fine classification that is difficult to evaluate with adequacy due to a difference in a single word or phrase, but it has still a limitation in that it cannot represent the performance of the model as a quantified numerical value.

## 3 Suggested Model

Sentences of the machine translation model can be translated using different vocabulary and sentence structures while having the same meaning as the correct sentences. However, BLEU commonly used today cannot evaluate lexical and structural significance because it compares the output sentences of machine translation only within the fixed correct sentences to measure the score and a lower score will be given if the translated sentences do not match the string in the correct sentences. Besides, the existing qualitative evaluations have a problem that the score cannot be quantified or varies depending on the evaluator.

This work proposes Grammar Accuracy Evaluation (GAE), a qualitative evaluation that can quantify and evaluate the grammatical significance of translated sentences to overcome the problems of today's methods of measuring the performance of machine translation models. GAE consists of 9 measurement categories that can be objectified by clear criteria - articles (in Korean, postpositional particle), vocabulary selection, singular or plural, misspelled words, missing words, added words, word order, tense, and sentence structure - and 0 (flawed) or 1 (not flawed) is assigned according to the presence or absence of flaws for each category.

The criteria for checking whether each category is flawed are as follows. "Article (or postpositional particle)" category evaluates omissions and grammatical flaws of articles (or postpositional particles). "Vocabulary selection" category evaluates whether there is a flaw according to the appropriateness of the sentence meaning determined by the syntactic composition of individual words as well as the meaning of individual vocabulary at the word level. "Singular or plural" category evaluates the number concordance of nouns, and it evaluates whether a noun specified in singular or plural in the source sentence expresses the same number in the target sentence. "Misspelled words" category evaluates whether misspelled spellings are added, or some spellings are omitted. "Missing words" category evaluates a case in which a vocabulary corresponding to the expression in the original text is omitted, and "added words" evaluates a case in which an expression not in the original text is added in the translation. "Word order" evaluates whether the translated sentence is correctly described according to the syntactic word order rule of the target language. "Tense" category evaluates whether the tense in the original text is expressed in the same tense in the translated text. "Sentence structure" category evaluates whether the sentence structure of the original text and the translated text match at the clause level.

| Sentence | Evaluation category | | | | | | | | | |
|---|---|---|---|---|---|---|---|---|---|---|
| | Article (en, de)/ Particle (ko) | Vocabulary selection | Singular/ Plural | Misspelled word | Missing word | Added word | Word order | Tense | Sentence structure | Sentence score |
| 1 | 1 | 1 | 1 | 1 | 1 | 1 | 1 | 1 | 1 | 100 |
| 2 | 1 | 1 | 1 | 1 | 1 | 1 | 1 | 1 | 1 | 100 |
| 3 | 1 | 0 | 1 | 0 | 1 | 1 | 1 | 1 | 0 | 66.66 |
| 4 | 1 | 0 | 1 | 1 | 0 | 1 | 1 | 1 | 1 | 77.78 |
| 5 | 1 | 1 | 1 | 0 | 1 | 1 | 1 | 1 | 1 | 88.89 |
| 6 | 1 | 0 | 1 | 1 | 1 | 1 | 1 | 1 | 1 | 88.89 |
| 7 | 1 | 0 | 1 | 1 | 0 | 1 | 1 | 1 | 1 | 77.78 |
| 8 | 1 | 1 | 1 | 1 | 1 | 0 | 1 | 1 | 1 | 88.89 |
| 9 | 1 | 0 | 1 | 1 | 1 | 1 | 1 | 1 | 1 | 88.89 |
| 10 | 1 | 0 | 1 | 1 | 0 | 1 | 1 | 1 | 1 | 77.78 |
| Category score | 100 | 40 | 100 | 80 | 70 | 90 | 100 | 100 | 90 | 85.56 (Model score) |

Table 1: Example of GAE score calculation table

An example of the measurement method is shown in Table 1. Each sentence score is calculated by expressing the average (0~1) of the scores for each category as a percentage between 0 and 100.

The score for each category of the entire sentences calculated in the same way indicates the accuracy of the specific grammar of the model and the average of all category scores is the GAE score, which represents the grammatical significance of the model. In this work, 50 translated sentences from the test data not included in model training were obtained from each model, and then the sentences were evaluated according to the GAE assessment categories.

## 4 Experiment

### 4.1 Experimental data and methods

In this work, we used two types of corpora for our experiment. First, the Korean-English parallel news corpus provided by AI Hub was used to train the Korean-English and English-Korean machine translation models, and it consists of 800,000 pairs of sentences in total. Next, Europarl v10 and News Commentary v15 English-German parallel corpus provided by WMT20 were used to train the English-German and German-English machine translation models. By selecting 350,000 sentences from the Europarl v10 corpus and 450,000 sentences from the News Commentary v15 corpus, 800,000 pairs of English-German parallel corpuses with the same number as the Korean-English parallel corpus were constructed. The two parallel corpora are randomly divided at a ratio of 98:1:1, respectively, and reconstructed into training data (784,000 sentences), validation data (1,000 sentences), and test data (1,000 sentences).

To build the experimental model, OpenNMT-py[4] was used, which is an open source library for machine translation model training developed by Harvard NLP group and SYSTRAN in December 2016. The Korean-English, English-Korean, German-English, and English-German models were trained by repeating 100,000 epochs, respectively. The hyper parameters applied to the model construction were set as follows. The encoder and decoder types were set to Transformer [5], the number of layers of the encoder and decoder is 6, the output dimension of encoder and decoder is 512, the number of hidden layers in the recurrent neural network is 512, the number of layers of the feedforward network is 2048, the number of heads of multi-head attention is 8, and the number of batch sizes is 4096. The dropout was set to 0.1, and the Adam optimization function was used to train the models under the same conditions. For tokenization, BPE[6] was applied, and the number of tokens used was set equally to 32,000 for all training models.

## 4.2 Result

### 4.2.1 Analysis of BLEU Scores

| Epoch | KO-EN | EN-KO | DE-EN | EN-DE |
|---|---|---|---|---|
| 10,000 | 23.35 | 8.83 | 27.25 | 20.98 |
| 20,000 | 28.19 | 11.96 | 29.74 | 23.83 |
| 30,000 | 29.80 | 12.63 | 30.42 | 24.13 |
| 40,000 | 30.09 | 12.81 | 30.59 | 24.41 |
| 50,000 | 30.71 | 13.06 | 30.64 | 24.58 |
| 60,000 | 30.40 | 13.21 | 30.61 | **24.63** |
| 70,000 | 30.39 | 13.16 | **30.73** | 24.24 |
| 80,000 | 30.50 | 13.28 | 30.66 | 24.33 |
| 90,000 | 30.80 | **13.35** | 30.47 | 24.18 |
| 100,000 | **30.93** | 13.29 | 30.26 | 24.02 |

Table 2: BLEU scores by model according to epoch change

Table 2 shows the results of measuring BLEU scores of four machine translation models, Korean-English, English-Korean, German-English, and English-German, every 10,000 epochs. As a result of measuring the performance of the models with BLEU, the Korean-English model had the highest score with 30.93, followed by the German-English model with 30.73, but the score difference between the two models was very insignificant at 0.2. The English-German model showed a score of 24.63, and the English-Korean model showed the lowest BLEU score with a score of 13.35.

As a result of the BLEU scores, it is confirmed that a higher score was measured when the target language was set to English than when the target language was set to Korean or German. The BLEU score of the Korean-English model was 149% higher than that of the English-Korean model, and the BLEU score of the German-English model was 25% higher than that of the English-German model.

### 4.2.2 Analysis of GAE Scores

| Category | KO-EN | EN-KO | DE-EN | EN-DE |
|---|---|---|---|---|
| Article (en, de)/ Particle(ko) | 84 | 96 | 98 | 86 |
| Vocabulary selection | 50 | 34 | 52 | 30 |
| Singular/ Plural | 82 | 100 | 96 | 96 |
| Misspelled word | 90 | 78 | 86 | 86 |
| Missing word | 66 | 72 | 72 | 70 |
| Added word | 82 | 82 | 84 | 74 |
| Word order | 94 | 92 | 94 | 94 |
| Tense | 98 | 98 | 98 | 92 |
| Sentence structure | 80 | 76 | 72 | 76 |
| **Model score** | **80.67** | **80.89** | **83.56** | **78.22** |

Table 3: GAE scores by model

Table 3 shows the GAE scores for each model measured according to 9 evaluation categories, and each GAE score was measured using the model at the epoch point showing the highest score based on the BLEU score.

In the GAE scores, calculated as the average of all category scores, the German-English model showed the highest score of 83.56, and the English-Korean model showed the next highest score with 80.89. Following that, the Korean-English model scored 80.67, and the English-German model showed the lowest score with 78.22. As a result of GAE score measurement, it is confirmed that all four experimental models show a score distribution within 78-84 points, and there is no significant difference in scores between the models.

The Korean-English model, which had the highest score when measured by BLEU, showed the third performance among the four models in the GAE scores, and the German-English model, which showed a BLEU score that was 0.2 lower than that of the Korean-English model, showed a rather high score of 2.67 in the GAE score. In the English-Korean model, the results of the GAE score and the BLEU score were the most contradictory. The BLEU score of this model was 13.35, the lowest among the four models, while the GAE score was 80.89, 0.22 higher than the Korean-English model, which had the highest BLEU score, showing the second highest score.

| Original | It is said they also **use relatively expensive** automobiles brands. |
|---|---|
| Answer | 자동차도 비교적 고가 브랜드를 이용하고 있다는 전언이다. |
| Translated | 상대적으로 가격이 비싼 자동차 브랜드도 사용한다고 한다. |

Table 4: Example of evaluation sentences of the English-Korean machine translation model

Table 4 shows the original English sentence used in the English-Korean translation test, the correct answer sentence in Korean, and the Korean sentence translated by the machine translation model. As a result of measuring the GAE score by comparing the original sentence and the translated sentence according to 9 measurement categories, the above translated sentence showed correct grammaticality in all categories. However, when this sentence was evaluated according to the BLEU standard, a score of 0 was measured, confirming a contrasting evaluation score between the two evaluation methods.

The reason why the low BLEU score was measured is because the BLEU evaluated the words replaced with synonyms such as "상대적"–"비교적", "고가"–"가격이 비싼", "이용하다"–"사용하다" and sentences expressed in different sentence structures such as "자동차도 비교적 고가 브랜드를 이용하고 있다"–"상대적으로 가격이 비싼 자동차 브랜드도 사용한다" as incorrect answers. The cause of the low BLEU score lies in the limitations of the N-gram-based BLEU measurement.

This problem is also confirmed in the English-German translation sentence in Table 5. This translated sentence showed correct grammaticality in all categories of GAE, while the BLEU score was 0, showing a contrasting result. "Experienced a decline" in the original English sentence presented as "eingeschränkt worden" in the German sentence was translated as "verzeichneten einen Rückgang" in the German translation sentence, and the English "more than" presented as "über" in the German sentence was translated as "mehr als" in the German translation. This shows that a low BLEU score is calculated due to the use of a different vocabulary and a change in the sentence structure even though the translation sentence is grammatically flawless.

| Original | More than 70 countries **experienced a decline** in freedom. |
|---|---|
| Answer | In über 70 Ländern ist die Freiheit **eingeschränkt worden**. |
| Translated | Mehr als 70 Länder **verzeichneten einen Rückgang** der Freiheit. |

Table 5: Example of evaluation sentences of the English-German machine translation model

By comparing the scores of each GAE category of the models, it can be confirmed that the linguistic characteristics between the source and target language have a significant effect on the translation result. As for the category score of "article (or postpositional particle)", the German-English model showed the best accuracy with 98 points, and the English-German model trained by switching the source and target language with the same corpus showed 86 points. Although both English and German are languages with articles, the articles (definite article and indefinite article) in English are subdivided into 16 variants in German according to the characteristics of the gender, number, and case, respectively. Therefore, it can be seen that prediction was easily possible when moving from a complex feature to a simple feature, such as a German-English translation, but low accuracy was measured when moving from a simple feature to a complex feature, such as an English-German translation. Following this, the "single-plural" category of the English-Korean model scored 100 points, showing excellent translation results without grammatical flaws in the sample sentences. This is due to the grammatical characteristics that Korean often refers to plurals collectively as singular, and plural notation for individual nouns is not obligatory. On the contrary, the English-Korean model showed a relatively low score of 82 points, and it was confirmed that the characteristic of the plurality, which was not clearly specified in Korean, caused a high grammatical error compared to other models. "Vocabulary selection" category, which is of great importance in translation work, scored the lowest in all models. The German-English model scored 52 points, the Korean-English model scored 50 points, the English-Korean model scored 34 points, and the English-German model scored 30 points. This shows that fastidious vocabulary selection is required in translation in the order of German, Korean, and English. In the category of "misspelled words", the Korean-English model showed the lowest typographical error with 90 points, and the English-Korean model showed the most typographical error with 78 points. Furthermore, in all machine translation models, the case where original words were omitted in the translated sentences showed a score of 66 to 72, whereas the case where words not in the original sentences were added in the translation results showed a higher score of 74 to 84. It shows that the vocabulary omission error occurs more frequently than the vocabulary addition error.

## 5   Conclusion

In this work, we analyzed the performance of machine translation models with BLEU, which is widely used for quantitative evaluation, and the Grammatical Accuracy Evaluation (GAE), which is a qualitative evaluation proposed in this work. For the experiment, four types of machine translation models were trained: Korean-English, English-Korean, German-English, English-German, while switching the source and target language using the Korean-English and German-English corpora.

As a result of the experiment, the BLEU scores showed a significant score difference between the models, but as a result of analyzing the translated sentences using GAE, it was confirmed that the BLEU score difference does not represent a difference in actual performance. As a result of cross-validation of the GAE evaluation for sentences with BLEU score of 0, it was confirmed that the translated sentence delivered the same meaning and there were no grammatical defects even though the translated sentence was translated into a different vocabulary and composition from the target sentence. This is due to the intrinsic characteristics of natural language, in which there are multiple variants of sentences and vocabulary with the same meaning. The GAE metric makes it possible to evaluate the machine translation model qualitatively by supplementing the problems caused by the structural limitations of the existing quantitative evaluation and presenting a qualitative evaluation standard that can be quantitatively measured. These experimental results suggest that the completeness of a translated sentence cannot be evaluated because it is limited to a specific sentence, and a qualitative evaluation method that compares it with the actual translated sentence and analyzes the characteristics of errors in evaluating the machine translation model should be used done in parallel.

However, BLEU and GAE analyze model performance in different aspects and have different advantages and disadvantages. BLEU allows instant score measurement by a pre-designed algorithm but does not represent the absolute performance of the translation results. On the other hand, since the GAE score is evaluated by a metric based on the grammar of the translated sentences, it is possible to evaluate the specific quality of the translation results, while it requires the intervention of a person who directly inspects the sentences, and the performance of the model must be measured with limited sample sentences.

These characteristics show that the two metrics can complement each other and that both metrics are required to equally evaluate the quantitative and qualitative aspects of the machine translation model. In addition, if a function that inevitably requires human intervention due to the nature of qualitative evaluation is implemented as an automated algorithm, it is judged that it will become a quantitative qualitative evaluation metric that can replace the quantitative evaluation method of the current machine translation model. Therefore, it is considered that a follow-up study on an evaluation system that can automatically perform the current evaluation items is necessary.